\pdfoutput=1
\documentclass[11pt]{article}

\usepackage[preprint]{acl}

\usepackage{times}
\usepackage{latexsym}

\usepackage[T1]{fontenc}
\usepackage[utf8]{inputenc}

\usepackage{microtype}

\usepackage{inconsolata}

\usepackage{graphicx}

\usepackage{amsmath}

\usepackage{booktabs} % For formal tables
\usepackage{tabularx} % Allows for "X" column type
\usepackage{colortbl}
\usepackage{adjustbox}

\usepackage{pdfpages}
\usepackage{subcaption}

\usepackage{soul}
\usepackage{xcolor}
\usepackage{xspace}
\sethlcolor{yellow}
\newcommand{\gray}[1]{\textcolor[RGB]{128,128,128}{#1}}

\newcommand{\para}[1]{{\vspace{2pt} \noindent \textbf{#1}
    \hspace{6pt}}}
\newcommand{\OurMethod}{ScalingFilter\xspace}

\title{ScalingFilter: Assessing Data Quality through Inverse Utilization of \\ Scaling Laws}

\author{
 \textbf{Ruihang Li\textsuperscript{1,2}},
 \textbf{Yixuan Wei\textsuperscript{2,3}},
 \textbf{Miaosen Zhang\textsuperscript{2,4}},\\
 \textbf{Nenghai Yu\textsuperscript{1}},
 \textbf{Han Hu\textsuperscript{2}},
 \textbf{Houwen Peng\textsuperscript{2}\thanks{Corresponding author.}}
\\
 \textsuperscript{1}University of Science and Technology of China,
 \textsuperscript{2}Microsoft Research Asia, \\
 \textsuperscript{3}Tsinghua University,
 \textsuperscript{4}Southeast University \\
}

\begin{document}
\maketitle
\begin{abstract}
High-quality data is crucial for the pre-training performance of large language models. Unfortunately, existing quality filtering methods rely on a known high-quality dataset as reference, which can introduce potential bias and compromise diversity. In this paper, we propose \OurMethod, a novel approach that evaluates text quality based on the perplexity difference between two language models trained on the same data, thereby eliminating the influence of the reference dataset in the filtering process. An theoretical analysis shows that \OurMethod is equivalent to an inverse utilization of scaling laws. Through training models with 1.3B parameters on the same data source processed by various quality filters, we find \OurMethod can improve zero-shot performance of pre-trained models in downstream tasks. To assess the bias introduced by quality filtering, we introduce \textit{semantic diversity}, a metric of utilizing text embedding models for semantic representations. Extensive experiments reveal that \textit{semantic diversity} is a reliable indicator of dataset diversity, and \OurMethod achieves an optimal balance between downstream performance and semantic diversity.
\end{abstract}

\section{Introduction}

The success of large language models (LLMs) is significantly influenced by the quality and quantity of the pre-training corpus. Researchers have developed various data curation pipelines to enhance dataset quality, focusing on raw web crawling, text extraction, repetition and toxic content removal, and, notably, quality filtering~\citep{gpt3, gopher, refinedweb}. 

\begin{figure}[tbp]
\centering
\vspace{-3mm}
\includegraphics[width=0.48\textwidth]{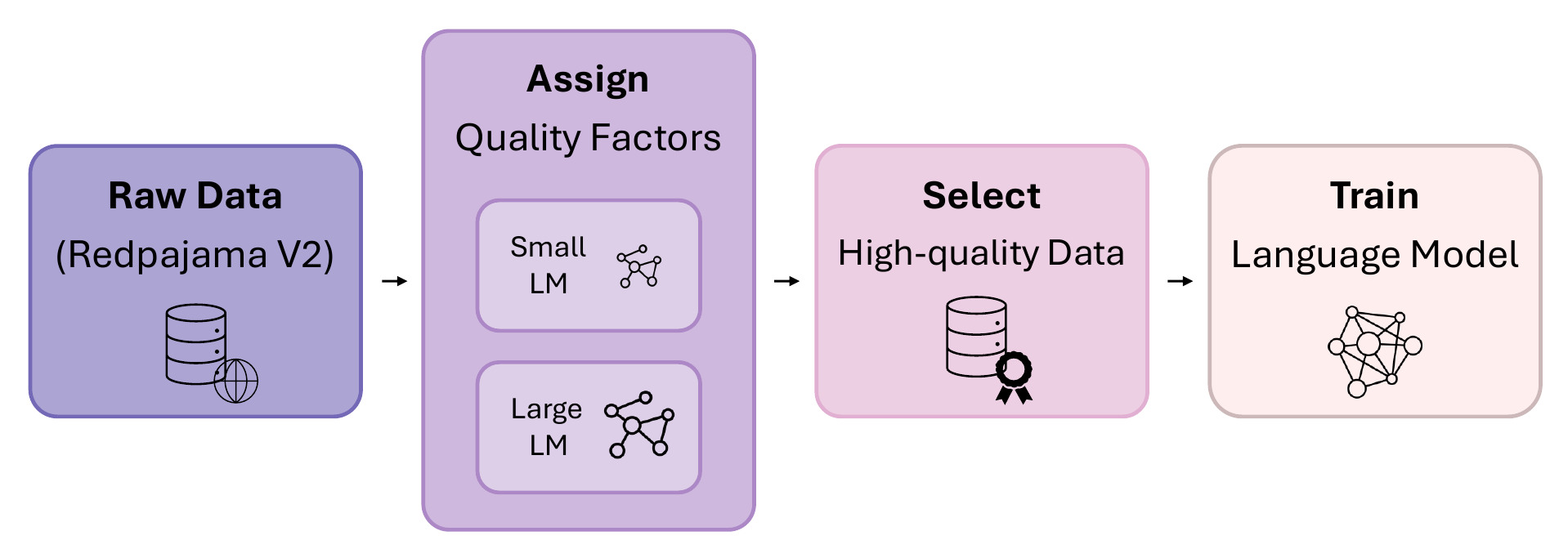}
\caption{In \OurMethod, we assess the quality of text documents by their scaling characteristics with language models in different sizes.}
\vspace{-5mm}
\label{fig:system-design}
\end{figure}

Quality filters aim to extract high-quality data from a noisy raw corpus, thereby improving the language model's performance without increasing training costs.
Existing filters are broadly classified into two categories: reference-dependent and reference-free approaches. Reference-dependent methods, such as binary classification~\citep{gpt3, pile, palm} and DSIR~\citep{dsir}, filter out low-quality data by comparing it with high-quality seed datasets. While effective, these methods inevitably introduce biases present in the reference data, such as specific writing styles or topics, thereby limiting the diversity and representativeness of training corpus~\citep{dolma}. In contrast, reference-free methods, such as perplexity gating~\citep{data-pruning}, assess data quality using predefined metrics like perplexity scores from pre-trained models. These methods mitigate the biases introduced by reference datasets but encounter challenges due to the indirect relationship between absolute perplexity and document quality. This  indirect relationship inadvertently favor data with simple and repetitive content. 
Although such content is easier for models to predict, it contributes minimally to learning diversity and complexity~\citep{qurating}.

To address these issues, we introduce a simple yet effective quality filtering approach named \OurMethod, which inversely leverages recent scaling laws in generative modeling to assess data quality. The core idea is to analyze the perplexity differences between two pre-trained models on the same data and assess the data quality based on these differences. We find a positive correlation between data quality and perplexity differences by inversely deriving Chinchilla scaling law \cite{chinchilla}. In other words, given a pair of pre-trained models of different sizes, documents with higher perplexity differences indicate higher quality.

\OurMethod involves utilizing the difference between two separate models for data quality assessment, effectively addressing the bias issue induced by relying on a single model trained on the reference data. This approach also mitigates the problem of selecting simple and repetitive texts that arise from overfitting to the perplexity metric, thereby enhancing data diversity and complexity. Furthermore, \OurMethod offers a theoretical analysis for using perplexity differences as a quality indicator for data filtering by inversely deriving model scaling laws.

Our experiments demonstrate that \OurMethod is superior to existing methods in improving filtered data quality while preserving data diversity. Specifically, we employ a pair of meta-models with sizes of 124M and 774M parameters to assess the perplexity difference for each document in the raw dataset, and then select the high-quality ones using a top-$k$ strategy. 
We train a 1.3B model from scratch using filtered high-quality data. We then evaluate its zero-shot performance on downstream tasks and assess the semantic diversity of the filtered dataset.

The results demonstrate that \OurMethod outperforms the unfiltered baseline and previous state-of-the-art (SoTA) quality filtering methods. Specifically, compared to the unfiltered baseline, \OurMethod achieves a +3.09\% improvement in downstream accuracy and a +2.23 increase in semantic diversity. When compared with perplexity gating~\citep{data-pruning, ccnet}, \OurMethod achieves a +1.12\% improvement in performance and a +4.7 increase in semantic diversity.

In summary, the contributions of this work are threefold:
\vspace{-0.5em}
\begin{enumerate}
    \item We introduce \textit{quality factor}, a novel metric that correlates directly with the quality of training data through the lens of model scaling laws, offering a more precise and unbiased approach to data curation.
    \item We propose \OurMethod, a new quality filtering method that utilizes the \textit{quality factor} to curate high-quality datasets without relying on reference data, thereby mitigating the risk of bias and enhancing the representativeness of the training corpus.
    \item To evaluate the data diversity of filtered datasets, we introduce \textit{semantic diversity} as a novel and reliable metric. Extensive experiments demonstrate that \OurMethod\ more effectively preserves the richness and variety of the raw data compared to conventional quality filtering approaches.
\end{enumerate}

\begin{figure*}[t]
    \vspace{-5mm}
    \centering
    \begin{subfigure}[b]{0.45\textwidth}
        \centering
        \includegraphics[width=\textwidth]{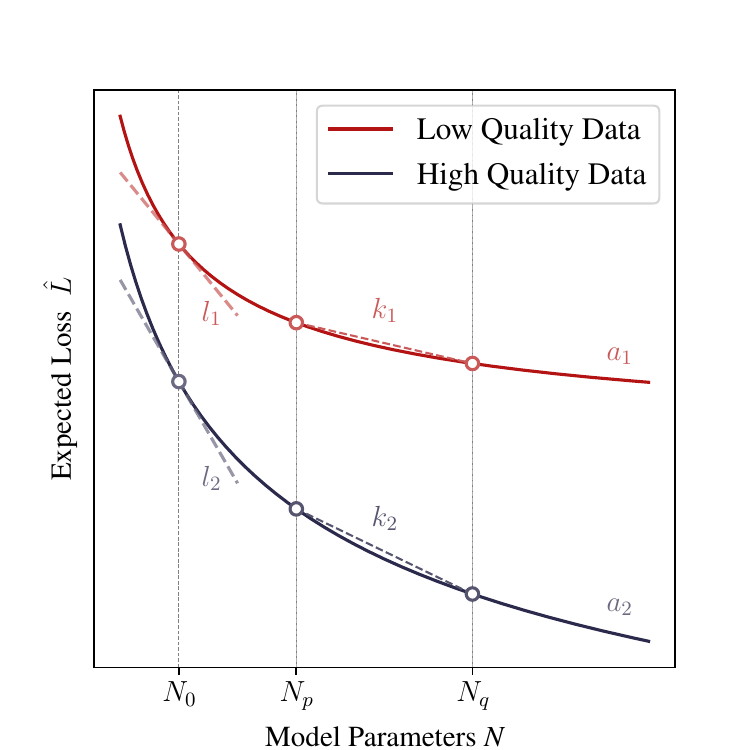}
        \caption{}
        \label{fig:method-illu}
    \end{subfigure}
    \hfill
    \begin{subfigure}[b]{0.45\textwidth}
        \centering
        \includegraphics[width=\textwidth]{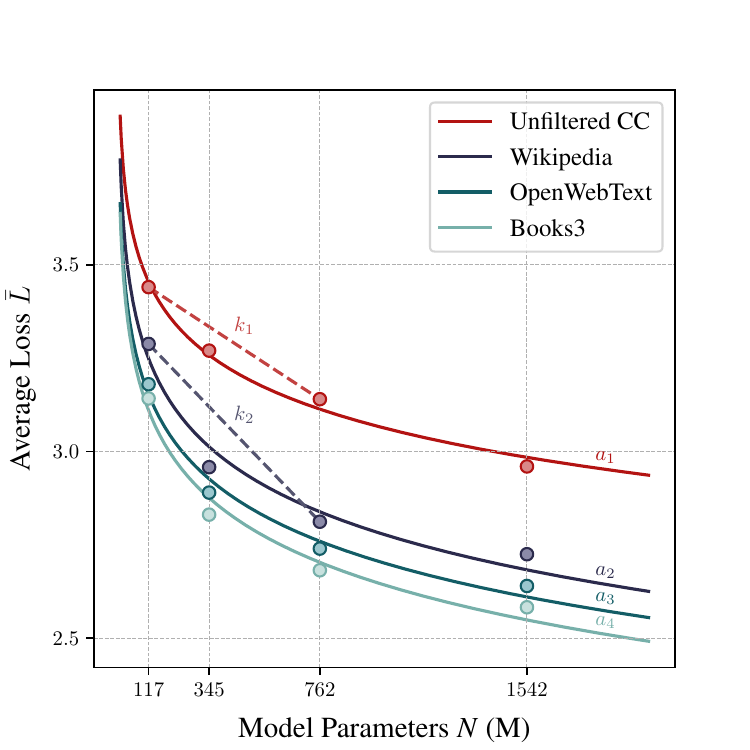}
        \caption{}
        \label{fig:method-avg}
    \end{subfigure}
    \vspace{-2mm}
    \caption{
    \textbf{(a)} A visual diagram illustrates the theoretical result that high-quality data accelerates the rate of loss decrease as model parameters increase, resulting in larger model scaling exponents $a$.
    \textbf{(b)} We calculated the average loss of GPT-2 models of different sizes on several datasets with recognized quality levels: Wikipedia, OpenWebText, and Books3 represent high-quality data, while Unfiltered CommonCrawl represents low-quality data. The results closely align with the theoretical analysis shown in (a), which indicates that high-quality data accelerates the rate of loss decrease as model parameters increase.
    }
    \vspace{-2mm}
    \label{fig:method-full}
\end{figure*}

\section{Methodology}
\label{sec:method}

\para{Overview.} Existing quality filtering methods depend on either a reference high-quality dataset or the absolute perplexity scores of documents from a single language model, which can introduce potential bias or result in inferior performance. In this section, we will elaborate on the principles of \OurMethod through mathematical derivation. The core concept of \OurMethod lies in estimating the quality of data samples by inversely applying the scaling law. Specifically, the scaling law reveals a power-law decrease in loss with increasing model size or data size~\citep{2017scalinglaw, openai-scalinglaw, chinchilla, meta-mm-scalinglaw}. Ultimately, the scaling law yields the optimal model/data scaling-up allocation strategy. In other words, under the same computational budget (TFLOPS), it determines the optimal ratio of model size to the number of training tokens to achieve the lowest loss, represented by a model scaling exponent $a$ and a data scaling exponent $b$.

Extensive experiments comparing multiple datasets with known quality differences revealed that high-quality data increases the model scaling exponent $a$~\citep{deepseekllm}. 
Specifically, the experiments compared the early and final versions of in-house data together with OpenWebText2~\citep{pile}, revealing that the final version of OpenWebText2 results in the highest $a$, while the early version with the poorest quality leads to the lowest $a$. 
Intuitively, a higher value of $a$ accelerates the rate at which the loss decreases as the model parameters increase. This positive relationship will be demonstrated later.
Such an observation suggests that high-quality data enhances logical clarity and decreases predictive difficulty after adequate training. Consequently, scaling up the model size becomes more beneficial when the compute budget is increased~\citep{deepseekllm, meta-mm-scalinglaw, openai-scalinglaw, chinchilla}. By inversely applying this principle, \OurMethod estimates data quality based on the rate of loss decrease in models with a fixed parameter difference, thereby separating high-quality data from the raw dataset.

To proceed, we will first define the \textit{quality factor}, which is the magnitude of loss reduction. Then, starting from the formula of the scaling law, we will demonstrate the positive correlation between the \textit{quality factor} and the model scaling exponent $a$. Furthermore, based on the positive correlation between $a$ and data quality observed in \citep{deepseekllm}, we can ultimately prove the positive correlation between the \textit{quality factor} and data quality.

\para{Quality factor.} We begin with defining the \textit{quality factor}, which we will later demonstrate to have a positive correlation with data quality. We denote the smaller meta-model as $p$ and the larger one as $q$. Both meta-models share the same architecture and are trained on the same dataset, with the only difference being the parameter counts: $N_p$ for $p$ and $N_q$ for $q$, with $N_p < N_q$. Let $x_i$ be a given text sample, and denote the \textit{quality factor} of this sample as $d_i$. Then, we have:

\begin{equation}
\label{eq:quality-factorx}
    d_i := \frac{\mathrm{PPL}_p (x_i)}{\mathrm{PPL}_q (x_i)}
\end{equation}

\noindent where $\mathrm{PPL}_p (x_i)$ and $\mathrm{PPL}_q (x_i)$ represent the perplexity scores of the text sample $x_i$ when evaluated by $p$ and $q$, respectively. It's important to note that perplexity has a direct relationship with the cross-entropy loss $L$ because $\mathrm{PPL} = 2^{L}$, indicating that the perplexity score is positively related to the loss $L$.

\para{Quality factor is positively correlated with data quality.} Next, we will introduce the expression of the scaling law~\citep{chinchilla, openai-scalinglaw, openai-mm-scalinglaw, meta-mm-scalinglaw} and transform it into a form involving the model scaling exponent $a$ which, as we introduced in the overview, is known to have a positive relationship with data quality~\citep{deepseekllm}. 

Given the number of model parameters $N$ and training tokens $D$, the expected model loss $\hat{L}$ is formulated~\citep{chinchilla} as:

\begin{equation}
    \hat{L} (N, D) = E + \frac{A}{N^\alpha} + \frac{B}{D^\beta}
\end{equation}

\noindent where $E$ represents the minimal achievable loss, corresponding to the entropy of natural text. The terms $\frac{A}{N^\alpha}$ and $\frac{B}{D^\beta}$ account for the functional approximation error and the optimization or convergence error, respectively~\citep{meta-mm-scalinglaw}. Here, $A$, $B$, $\alpha$, and $\beta$ are hyperparameters related to the model architecture and the training data. The scaling law, indicating the optimized numbers of $N$ and $D$ under a given compute budget $C$, follows a power-law form~\citep{openai-scalinglaw, chinchilla}:

\begin{equation}
    N_{opt}\propto C^{a}\quad D_{opt}\propto C^{b}
\end{equation}

\noindent where $a=\frac{\beta}{\alpha+\beta}$ and $b=\frac{\alpha}{\alpha+\beta}$ represent the model and data scaling exponents~\citep{deepseekllm, chinchilla} respectively, indicating the proportions of the total computational budget allocated to model scaling and data scaling in the optimal computation allocation.

Consider setting $\eta \doteq \alpha + \beta$, then $\hat{L}$ can be presented as:

\begin{equation}
    \hat{L} (N, D) = E + \frac{A}{N^{(1-a)\eta}} + \frac{B}{D^{a\eta}}\  
\end{equation}

\noindent We focus on the relationship between expected loss $\hat{L}$ and model scaling exponent $a$ as well as model size $N$, and thus denote $\hat{L}$ as $\hat{L}(a, N)$. It's obvious that $\hat{L}$ decreases as $N$ increases. We further prove in Appendix~\ref{app:dv1} that at a specific $N_0$, the slope of the tangent to the $\hat{L}-N$ curve decreases as $a$ increases (i.e., the larger the $a$, the steeper the tangent, as illustrated in Figure~\ref{fig:method-illu} that $l_2$ is steeper than $l_1$). Due to this monotonic relationship, we can infer the value of $a$ from the slope of the tangent: for a given $N_0$, a steeper tangent (greater absolute value of the slope) indicates a larger $a$, that is:

\begin{equation}
    a \propto -\frac{\partial L}{\partial N} \bigg|_{N = N_0}
\end{equation}

Furthermore, we prove in Appendix~\ref{app:dv2} that the above conclusion can be extended from the tangent slope at a given $N_0$ to the slope of the secant line for any given $\Delta N$ (i.e. $k_i$ in Figure~\ref{fig:method-illu}). Letting $\Delta N = N_q - N_p$, the slope of the secant line is always negative, and $a$ is positively correlated with the negative slope of the secant line. Since the quality factor $d$ is also positively correlated with the negative slope of the secant line, it follows that $d$ is positively correlated with $a$:

\vspace{-5mm}
\begin{equation}
\begin{array}{c}
\begin{cases}
    a \propto -\frac{\hat{L}(N_q) - \hat{L}(N_p)}{N_q - N_p}, \\[10pt]
    d = 2^{\hat{L}(N_p) - \hat{L}(N_q)} = 2^{-(\hat{L}(N_q) - \hat{L}(N_p))}
\end{cases} \\[20pt]
\implies d \propto a 
\end{array}
\end{equation}
Based on empirical observations from~\citep{deepseekllm}, higher values of $a$ are achieved with high-quality data, indicating that \textbf{\textit{quality factor} $d$ is positively correlated with data quality}. 

This conclusion aligns with our practical comparative tests. As shown in Figure~\ref{fig:method-avg}, we calculated the average loss of GPT-2 models of different sizes on several datasets with recognized quality levels: Wikipedia, OpenWebText, and Books3 represent high-quality data, while Unfiltered CommonCrawl represents low-quality data, based on a random sample of 10k documents from each dataset. The results align closely with the theoretical estimates shown in Figure~\ref{fig:method-illu}, where the high-quality data shows a steeper secant ($k_2 > k_1$) compared to low-quality data. It's worth noting that a single case might deviate from the training data distribution of the meta-models, leading to higher absolute perplexity for various model sizes. Thus, relying solely on single perplexity as a quality criterion can result in misjudgments. However, high-quality data follows a law where perplexity decreases more with an increase in model parameters, indicating a greater perplexity difference (i.e., \textit{quality factor}).

\para{Selecting high-quality data with \textit{quality factor}.} 
We have demonstrated that the \textit{quality factor} can directly characterize data quality above, so it's straightforward to directly use it to select high-quality data from a noisy unfiltered dataset. We call this simple yet effective method as \textbf{\OurMethod}, as illustrated in Figure~\ref{fig:system-design}. Consider a unfiltered set of documents $\mathcal{S}$, containing both high and low-quality documents. For each sample $x_i \in \mathcal{S}$, we calculate the quality factor $d_i$ for it. As derived previously, samples with higher $d_i$ are of better quality. The top-$k$ samples are then selected based on the desired cleaning rate to form the resulting dataset.

\section{Experiments}
\label{sec-exp}

In this section, we will demonstrate the effectiveness of \OurMethod through extensive experiments. Specifically, language models trained on data filtered by \OurMethod consistently achieved superior performance across various downstream tasks, compared to the unfiltered baseline and other common quality filtering approaches, highlighting the higher quality of the data. Furthermore, by measuring the semantic diversity of the filtered dataset, we found that \OurMethod effectively preserved the diversity present in the original dataset.

\begin{table*}[!htbp]
\centering
\caption{
Zero-shot downstream accuracy of models trained with different quality filters. We cover a variety of tasks and widely used datasets~\citep{refinedweb, gpt3, palm, cerebras, pythia}, including sentence completion, coreference resolution, natural language inference and multiple-choice question answering. For binary classification \cite{gpt3, palm, llama} and importance resampling~\citep{dsir}, we leverage the best results from various reference datasets, whereas perplexity gating~\citep{data-pruning} utilizes the larger model's perplexity in our meta-models.
}
\label{tab:vs-quality-filter}
\begin{adjustbox}{max width=\textwidth}
    \begin{tabular}{lcccccccc}
    \hline
    Quality Filter & Hellaswag & LAMBADA & Winogrande & PIQA & ARC & OpenbookQA & BoolQ & Avg\\ \hline
    Random & 45.40 & 41.96 & 51.07 & 69.80 & 39.88 & 32.40 & 56.76 & 48.18 \\
    Binary Classification & 48.13 & \textbf{48.96} & 54.30 & 69.75 & 41.66 & 30.40 & 61.38 & 50.65 \\
    Importance Resampling & 47.52 & 48.36 & 54.38 & 68.50 & 41.63 & \textbf{32.60} & 60.80 & 50.54 \\
    Perplexity Gating & 48.17 & \textbf{48.96} & 53.04 & 69.75 & 41.54 & 29.60 & 60.00 & 50.15 \\
    \textbf{\OurMethod (Ours)} & \textbf{49.07} & 48.42 & \textbf{55.09} & \textbf{70.57} & \textbf{42.67} & 31.40 & \textbf{61.68} & \textbf{51.27} \\ \hline
    \end{tabular}
\end{adjustbox}
\vspace{-3mm}
\end{table*}

\subsection{Data Quality Evaluation}
\label{sec:downstream}

\para{Setup.} We begin with five CommonCrawl dumps from 2019 to 2023, processed through the CCNet~\citep{ccnet} pipeline, in accordance with~\citep{redpajamav1}. From the preprocessed dataset, 500 GB of text data are randomly selected as our baseline, yielding approximately 125 billion tokens for additional quality filtering. In each experiment, we train a decoder-only model with 1.3B parameters, using the same model architecture as~\citep{fp8lm}. Each model is trained on 25B tokens until performance levels off, according to~\citep{chinchilla, refinedweb, data-pruning}, which takes approximately 4 days on 4 nodes with 8 NVIDIA Tesla V100 GPUs. We use pre-trained GPT-2 models~\citep{gpt2} as default meta-models to calculate quality factors for each sample. The smaller and larger models have 124M and 774M parameters, respectively. We later perform ablation studies to discuss impacts by the pre-training data. Following~\citep{refinedweb}, we utilize the \texttt{lm-evaluation-harness} library~\citep{evalharness} to evaluate zero-shot performance across various downstream tasks of each model trained on documents retained through specific quality filtering method. We encompasses a variety of tasks and widely used datasets~\citep{refinedweb, gpt3, palm, cerebras, pythia}, including sentence completion (Hellaswag~\citep{hellaswag} and LAMBADA~\citep{lambada}), coreference resolution (Winogrande~\citep{winogrande}), natural language inference (ARC~\citep{arc}), and multiple-choice question answering (PIQA~\citep{piqa}, OpenbookQA~\citep{openbookqa}, BoolQ~\citep{boolq}).

\para{Baselines.} We compare \OurMethod\ with random selection, binary classification~\citep{gpt3, pile, palm, llama}, importance resampling~\citep{dsir} and perplexity gating~\citep{data-pruning}. All quality filters will keep 70\% of the unfiltered documents in align with~\citep{redpajamav1}, if not specified otherwise. As for binary classification, we choose Wikipedia, books and OpenWebText as positive samples and unfiltered CommonCrawl documents as negative ones, following~\citep{glam, palm}. We set the shape parameter of Pareto distribution $\alpha = 9$, following~\citep{gpt3, pile, palm}. As to importance resampling, we follow the settings in~\citep{dsir}, with OpenWebText~\citep{openwebtext} as the target dataset. As for perplexity gating, we follow~\citep{data-pruning} as well as our cleaning ratio, keeping documents with perplexity ranging from 15th to 85th percentiles, resulting in keeping the middle 70\% documents of the unfiltered dataset. Perplexity is computed by the larger of the meta-models, the one with higher capacity and ability.

\para{Results.} Table~\ref{tab:vs-quality-filter} shows the comparison between various data quality filter methods. In summary:

\begin{itemize}
\item On average, \OurMethod\ shows a 0.62\% improvement over the widely-used binary classification quality filtering method and a 0.73\% improvement over importance resampling, achieving the state-of-the-art performance.
\item \OurMethod\ achieves a 1.12\% improvement in average accuracy over perplexity gating, a competing reference-free quality filtering approach.
\end{itemize}

Notably, for binary classification~\citep{gpt3, palm, llama} and importance resampling~\citep{dsir}, we use the best results from various reference datasets, specifically OpenWebText. The results for perplexity gating~\citep{data-pruning} use the larger model's perplexity in our meta-models for a fair comparison. Ablations concerning the reference datasets of the aforementioned methods will be discussed in subsequent sections.

\begin{table*}[!tbp]
\centering
\caption{
Ablations on effects of meta-models training data within the \OurMethod framework. The results reveal that meta-models trained on alternative datasets also showcase competitive performance, indicating that there is not an overly strong dependency on meta-models pretrained on WebText, emphasizing the robustness and flexibility of \OurMethod variants.
}
\label{tab:self-trained}
\begin{adjustbox}{max width=\textwidth}
    \begin{tabular}{lcccccccc}
    \hline
    Training Data & Hellaswag & LAMBADA & Winogrande & PIQA & ARC & OpenbookQA & BoolQ & Avg \\ \hline
    Unfiltered CC & 47.34 & 47.78 & 54.22 & \textbf{70.78} & 40.64 & 30.40 & 60.95 & 50.30 \\ 
    Wikipedia & 48.81 & 47.64 & \textbf{56.67} & 69.31 & 41.71 & \textbf{32.60} & 61.07 & 51.12 \\ 
    OpenWebText & 48.15 & 46.01 & 54.06 & 69.91 & 43.01 & 31.40 & 60.89 & 50.49 \\  
    WebText$^{\dagger}$ & \textbf{49.07} & \textbf{48.42} & 55.09 & 70.57 & \textbf{42.67} & 31.40 & \textbf{61.68} & \textbf{51.27} \\  \hline
    \end{tabular}
\end{adjustbox}
\raggedright
\scriptsize{$^{\dagger}$ For model pairs trained on WebText, we directly use OpenAI GPT-2 models from HuggingFace, which is the meta-models used in the original \OurMethod framework.}
\vspace{-3mm}
\end{table*}

\para{Meta-models trained on various datasets exhibit competitive and comparable performance.} We detail ablation studies with meta-models trained on different datasets in Table~\ref{tab:self-trained}. Besides the meta-models trained on WebText, results of which are shown in Table~\ref{tab:vs-quality-filter}, we trained meta-models on a subset of Wikipedia, OpenWebText, and unfiltered CommonCrawl data with no more than 25B tokens. Each dataset was used to train meta-models for 1 epoch. The results demonstrate that all experiments outperform the baseline of random selection. Training on unfiltered CommonCrawl or OpenWebText yielded results competitive with those from other quality filtering methods. Furthermore, training on Wikipedia achieved results very close to the best, with a marginal gap of 0.15\%.

\para{Ablations on different sizes of meta-models.} We perform experiments to investigate impacts brought by using pairs of meta-models with different sizes. The results are briefly presented in Table~\ref{tab:abla-sizes-detail}. When using a pair of meta-models with relatively small differences in the number of parameters to estimate the quality factors of data, there is a certain degree of performance degradation in the downstream tasks. Reducing the size of the larger models in meta-models from 774M to 335M decreases the average performance on downstream tasks by 0.96\%. Conversely, increasing the size of the smaller models in meta-models from 124M to 335M results in a decrease of 1.28\% in performance. This suggests that a larger parameter gap may more effectively amplify differences in how models fit textual data, allowing for a more reliable assessment of the quality factor. Detailed exploration of this hypothesis is left as future work.

\para{Ablations on reference datasets.} We also examine the impacts of different reference datasets on popular quality filtering methods that rely on a reference. Results are shown in Table~\ref{tab:abla-reference}. Binary classification using OpenWebText as the positive class results in the best performance, similar to importance resampling with the same dataset as a reference. This aligns with the findings in~\citep{deepseekllm}, which confirm that OpenWebText has superior data quality. Binary classification with a mixed dataset including OpenWebText, Wikipedia, and books yields inferior results, possibly due to the classifier's training recipe, such as the mixing ratio of the three datasets. Surprisingly, importance resampling with Wikipedia results in similar average accuracy to the random baseline, with much better accuracy in ARC and BoolQ but significantly worse performance in sentence completion tasks like Hellaswag and LAMBADA, possibly due to the serious domain shift towards Wikipedia. In conclusion, the choice of reference datasets has a significant impact on the performance of quality filters that rely on references.

\begin{table*}[!htbp]
\centering
\caption{
Ablation studies on the effects of reference data. We varied the reference datasets for binary classification~\citep{gpt3, palm, llama} and importance resampling~\citep{dsir}. The results indicate that OpenWebText is the optimal reference dataset choice for both reference-dependent quality filtering methods.
}
% \vspace{-2mm}
\label{tab:abla-reference}
\begin{adjustbox}{max width=\textwidth}
    \begin{tabular}{lcccccccc}
    \hline
    Quality Filter & Hellaswag & LAMBADA & Winogrande & PIQA & ARC & OpenbookQA & BoolQ & Avg \\ \hline
    Random & 45.40 & 41.96 & 51.07 & 69.80 & 39.88 & 32.40 & 56.76 & 48.18 \\ \hline
    \textit{Binary Classification} & & & & & & & & \\
    OpenWebText & \textbf{48.13} & \textbf{48.96} & 54.30 & \textbf{69.75} & 41.66 & 30.40 & 61.38 & \textbf{50.65} \\
    Wikipedia & 46.80 & 46.96 & 53.35 & 69.15 & 41.40 & 32.00 & 61.31 & 50.14 \\
    Mixed$^{\dagger}$ & 47.10 & 47.68 & 53.43 & 68.61 & 42.19 & 32.20 & 57.71 & 49.85 \\ \hline
    \textit{Importance Resampling} & & & & & & & & \\
    OpenWebText & 47.52 & 48.36 & \textbf{54.38} & 68.50 & 41.63 & \textbf{32.60} & 60.80 & 50.54 \\
    Wikipedia & 43.08 & 38.56 & 51.93 & 66.65 & \textbf{42.76} & 32.40 & \textbf{61.90} & 48.18 \\ \hline
    \end{tabular}
\end{adjustbox}
\raggedright
\scriptsize{$^{\dagger}$ This experiment uses a mixed dataset as reference dataset following~\citep{glam, palm}, with OpenWebText, Wikipedia, and books. The classification scores are directly obtained from quality signals provided by Redpajama V2.}
\vspace{-3mm}
\end{table*}

\subsection{Data Diversity Evaluation}
\label{sec:diversity}

Training large language models requires diverse data. Current quality filters, by favoring text data similar to the reference dataset, may discard documents on informal topics or from minorities, reducing the trained model's knowledge diversity~\citep{ccnet, dolma}. How can we assess a dataset's data diversity? We introduce a metric to measure a document group's semantic diversity. 

\para{Semantic diversity metric.} Following~\citep{vendi}, we define semantic diversity as the exponential of the Shannon entropy of the semantic similarity matrix's eigenvalues. For a set of text documents $x_1, x_2, ..., x_n$ and a semantic similarity function $f$, we obtain a similarity matrix $\mathbf{S}$, where each entry $s_{i, j} = f(x_i, x_j)$. Denoting $\lambda_1, \lambda_2, ..., \lambda_n$ as the eigenvalues of $\mathbf{S}/n$, we define semantic diversity as follows.

\vspace{-4mm}
\begin{equation}
\mathrm{Semantic Diversity} = \exp\left( - \sum_{i=1}^n \lambda_i \log \lambda_i \right)
\end{equation}

A pre-trained language model extracts each document's semantic embedding, using cosine similarity as the similarity function. In our experiments, we utilize the \texttt{bge-base-en-v1.5} model~\citep{bge} with the \texttt{sentence\_transformers} library due to its efficiency and outstanding performance in various text embedding-related retrieval and clustering tasks.

\begin{figure}[tbp]
\centering
\includegraphics[width=0.45\textwidth]{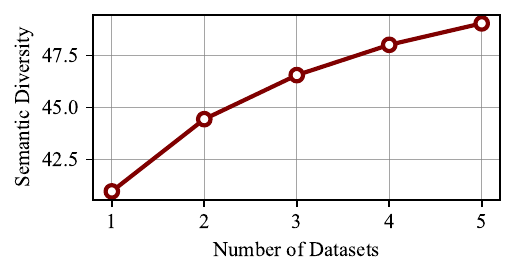}
\vspace{-2mm}
\caption{Positive correlation between the number of datasets and semantic diversity, demonstrating semantic diversity as a reliable measure of data diversity.}
\vspace{-6mm}
\label{fig:diversity-abla-work}
\end{figure}

\begin{figure}[htbp]
\centering
\includegraphics[width=0.5\textwidth]{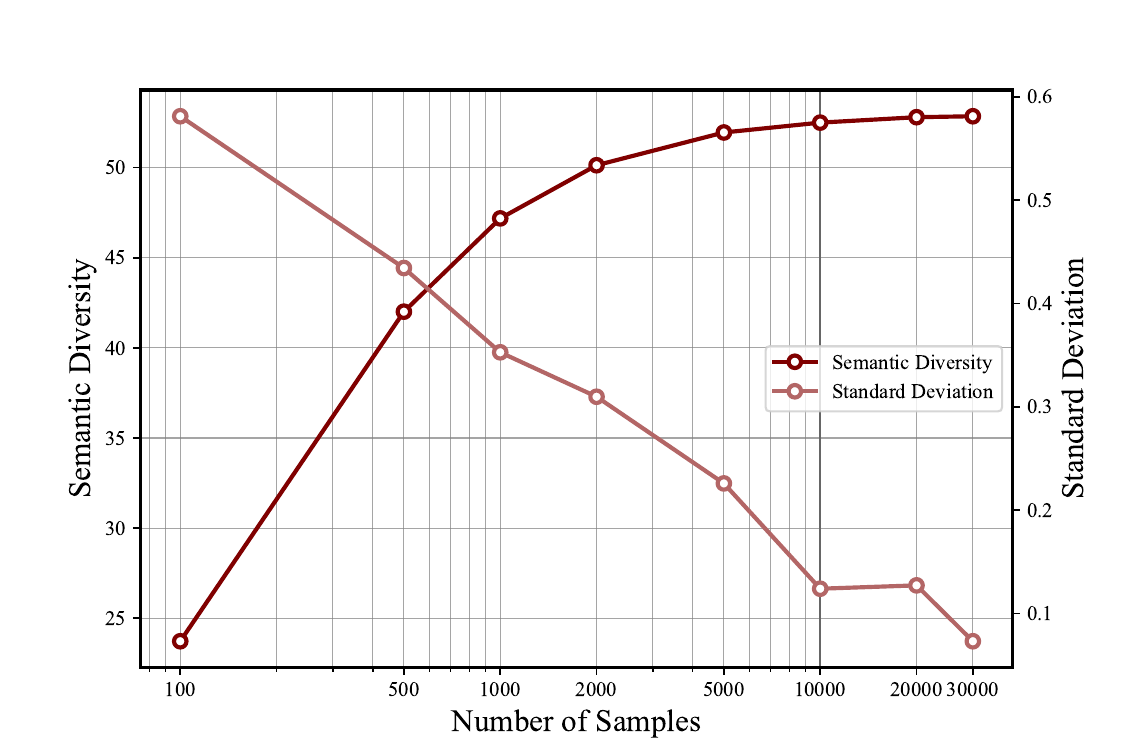}
\caption{Results on the relationship between semantic diversity and sample size. Semantic diversity stabilizes at a sample size of 10,000, with a standard deviation below 0.2. Therefore, we choose 10,000 as our sample size for calculating semantic diversity, as it represents the dataset's diversity adequately while ensuring computational efficiency.}
\vspace{-4mm}
\label{fig:diversity-abla-10ksamples}
\end{figure}

\para{Selecting a proper size of documents.} Computational constraints prevent calculating semantic diversity for all documents in the dataset. Experiments on the unfiltered dataset help select an appropriate document count for calculating the semantic diversity metric. For each experiment, we randomly select $n$ samples, calculate their semantic diversity score, and repeat this process 10 times to compute the average score and standard deviation. Results are displayed in Figure~\ref{fig:diversity-abla-10ksamples}. Results indicate that semantic diversity stabilizes when the group exceeds 10,000 samples, with a standard deviation of 0.12. We choose 10,000 samples for subsequent experiments to balance accuracy and efficiency.

\begin{table*}[!htbp]
\centering
\caption{
Ablations on effects of sizes of meta-models. To note, the original \OurMethod uses a pair of meta-models with 124M and 774M parameters, respectively.
}
\label{tab:abla-sizes-detail}
\vspace{-1mm}
\begin{adjustbox}{max width=\textwidth}
    \begin{tabular}{cccccccccc}
    \hline
    Small Model & Large Model & HellaSwag & LAMBADA & Winogrande & PIQA & ARC & OpenbookQA & BoolQ & Avg \\ \hline
    124M & 335M & 48.77 & 47.25 & 53.67 & 69.75 & 41.12 & \textbf{32.00} & 59.60 & 50.31 \\ 
    335M & 774M & 48.32 & 45.76 & 52.41 & 70.18 & 42.05 & 30.60 & 60.61 & 49.99 \\     
    124M & 774M & \textbf{49.07} & \textbf{48.42} & \textbf{55.09} & \textbf{70.57} & \textbf{42.67} & 31.40 & \textbf{61.68} & \textbf{51.27} \\  \hline
    \end{tabular}
\end{adjustbox}
\vspace{-3mm}
\end{table*}

\para{The proposed metric can reflect data semantic diversity.} Our experiments showed that semantic diversity effectively reflects data diversity under multi-datasets settings. We selected five datasets with diverse topics or writing styles, including news articles (CC-News~\citep{ccnews}), movie reviews (IMDB~\citep{imdb}), forums (Reddit \footnote{https://www.reddit.com}), Wikipedia, and crawled web pages (OpenWebText~\citep{openwebtext}). We extracted the same number of samples from one or more of the above datasets, creating a mixed subset of 10,000 samples. We then averaged the relationship between semantic diversity and the number of datasets (N). Figure~\ref{fig:diversity-abla-work} shows a positive correlation between semantic diversity and the number of datasets (N), indicating that semantic diversity accurately reflects data diversity within datasets.

\para{Quality filtering with quality factor keeps the diversity of the unfiltered dataset.} We assess the semantic diversity of datasets resulting from various quality filtering methods. The results are presented in Table~\ref{table:quality_diversity}.
Most quality filters achieve higher diversity than the original unfiltered dataset, likely due to the removal of a large number of machine-generated spams with similar semantic meanings. The results indicate that importance resampling achieves the highest diversity, at 56.25, attributed to its resampling strategy. \OurMethod results in greater diversity compared to the most commonly used binary classification, thanks to its reference-free nature. Perplexity gating reduces the diversity of the original dataset, supporting the conclusion from~\citep{ccnet} that filtering data based solely on perplexity thresholds can introduce unexpected bias to data diversity.

\begin{table}[ht]
\centering
{\fontsize{10pt}{13pt}\selectfont
\begin{tabular}{lc}
\hline
\textbf{Quality Filter}     & \textbf{Diversity} \\ \hline
Random                      & 52.50$_{0.12}$          \\ 
Binary Classification       & 53.99$_{0.19}$          \\ 
Importance Resampling       & \textbf{56.25$_{0.21}$} \\ 
Perplexity Gating           & 50.03$_{0.21}$          \\ 
\OurMethod                  & 54.73$_{0.14}$          \\ \hline
\end{tabular}
}
\caption{Quality filter methods and their impact on semantic diversity. The results represent averages over 10 attempts, with standard deviations noted as subscripts.}
\vspace{-4
mm}
\label{table:quality_diversity}
\end{table}

\section{Related Work}
% \para{Pipelines for pretraining data.} Training data for large language models typically comprises a curated dataset (e.g., Wikipedia, books~\citep{gutenberg}) and a scraped dataset (e.g., OpenWebText~\citep{openwebtext}, RefinedWeb~\citep{refinedweb}). Our focus here is on the scraped dataset. Scraped datasets often originate from large-scale automatic web crawls (e.g., CommonCrawl \footnote{https://commoncrawl.org}), and the quality of web content is highly variable. For instance, it may include low-quality data that can impair the performance of language models. Therefore, many practitioners construct a data preprocessing pipeline to enhance data quality~\citep{ccnet, refinedweb, redpajamav1}. Generally, these pipelines can be broken down into several steps~\citep{gpt3, gopher, llama}: (1) Text Extraction, extracting textual content from raw HTML; (2) Language Identification, filtering data in the desired language; (3) Deduplication, removing duplicate or similar documents; (4) toxic content \& PII removal, removing documents with blacklisted words or topics as well as removing personal identification information from the crawled web pages; (5) Quality Filtering, selecting high-quality documents out of noisy raw dataset. This work primarily focuses on the last step of the pipeline, namely quality filtering.

\para{Quality Filtering.} Pretraining data for large language models often includes low-quality content, such as harmful machine-generated spam or anomalous formats. To filter this data, researchers typically score documents using linear classifiers or language models, then filter based on these scores. High-quality data proxies like Wikipedia, books, and OpenWebText are commonly used. Early studies, such as~\citep{gpt3, palm}, employed linear binary classifiers, comparing curated datasets to unfiltered CommonCrawl data and used noisy thresholding with Pareto randomness, which, while potentially enhancing diversity, might reduce data quality as suggested by~\citep{dsir}. Recent studies, such as~\citep{llama}, use Wikipedia as the sole positive class and apply hard thresholding, potentially limiting corpus diversity and introducing biases. Another approach involves language models. For instance,~\citep{ccnet} trained an n-gram model on Wikipedia and categorized documents by perplexity into head, middle, and tail, with low-perplexity documents retained for pre-training~\citep{redpajamav1}. Other studies~\citep{ccnet, dolma} keep all data to preserve diverse writing styles. Similarly, some filter data based on a model's perplexity, which might bias towards easily predicted texts and discard challenging but high-quality content~\citep{data-pruning}. Our approach introduces a quality factor derived from two language models to address this issue.

\para{Scaling Laws.} Scaling laws quantify how model size, dataset size, compute budget, and performance relate during neural network training. Initial studies~\citep{2017scalinglaw, openai-scalinglaw} identified a power-law relationship among these factors.~\citep{chinchilla} introduced a unified formula for scaling laws that incorporates data-dependent scaling terms. Recent studies show variations in scaling laws across multilingual~\citep{crossling-scalinglaw} and multimodal~\citep{openai-mm-scalinglaw, cl-scalinglaw} settings.~\citep{meta-mm-scalinglaw} meticulously analyzed this subject, deriving varied scaling law parameters for uni-modal and mixed-modal contexts, highlighting the significant impact of data modality on scaling behaviors. This discovery suggests a hypothesis that varying data quality influences scaling behaviors. A recent study on large language model scaling laws~\citep{deepseekllm} confirms data quality impacts both model and data scaling exponents in scaling laws. This paper demonstrates the link between scaling law parameters and data quality, facilitating the selection of high-quality samples based on their scaling attributes.

\section{Conclusion}
\label{sec:conclusion}

We have presented \OurMethod for data quality filtering in a reference-free manner. Starting from the scaling law, we demonstrate that the perplexity difference across agnate models of different sizes (i.e. \textit{meta-models}) correlates with data quality positively. We select samples with higher perplexity difference (i.e. \textit{quality factors}) to form the pretraining dataset. By eliminating the bias brought by reference datasets, our method achieves better downstream performance over several strong baselines while preserving data diversity.

\section*{Limitations}

There are still some limitations of \OurMethod that need to be addressed. First, it relies on perplexity difference between two LLMs, which may miss nuanced aspects of text quality like factual accuracy or bias like race bias, social class bias and gender bias, etc. Second, it requires significant computational resources to compute perplexity differences for a large dataset. Third, its applicability to other languages and data-limited domains is uncertain. Future research should address these limitations and further explore the relationship between semantic diversity and model performance, particularly regarding fairness and bias.

\bibliography{custom}

\appendix

\renewcommand{\thetable}{A.\arabic{table}}
\setcounter{table}{0}

\section{Appendix}

\subsection{Hyperparameters of training language models}
\label{app:hyper}

We train decoder-only transformer~\citep{transformer} models using Megatron-DeepSpeed~\citep{mdeep, megatron}. The hyperparameters used in the training process is listed in Table~\ref{table:hyperpara-lm}.

\begin{table}[ht]
\centering
\begin{tabular}{lc}
\hline
$n$ params & 1.3B  \\ 
$n$ layers & 24 \\
$d$ model & 2048 \\
$n$ heads & 32 \\
$d$ head & 64 \\
Sequence length & 2048 \\
Global batch size & 256 \\
LR schedule & cosine \\
Learning rate & $2.5 \times 10^{-4}$ \\
Min LR & $2.5 \times 10^{-5}$ \\
Weight decay & 0.1 \\
Optimizer & Adam \\
Adam $\beta_1$ & 0.9 \\
Adam $\beta_2$ & 0.95 \\
Adam $\epsilon$ & $1.0 \times 10^{-8}$  \\
Tokenizer & cl100k\_base \\ \hline
\end{tabular}
\vspace{2mm}
\caption{Hyperparameters of training language models.}
\label{table:hyperpara-lm}
\end{table}

\subsection{Derivation of \OurMethod}
\label{app:derivation}

\subsubsection{Positive correlation between $a$ and the negative tangent slope}
\label{app:dv1}

Let's start with the parametric loss function introduced by Chinchilla~\citep{chinchilla} scaling law.

\begin{equation}
    \hat{L} (N, D) = E + \frac{A}{N^\alpha} + \frac{B}{D^\beta}
\end{equation}

\noindent where $E$ represents the minimal achievable loss, corresponding to the entropy of natural text. The scaling law, indicating the optimized numbers of $N$ and $D$, follows a power-law form:

\begin{equation}
    \begin{gathered}
        N_{opt}\propto C^{a}\quad D_{opt}\propto C^{b} \\
        \mathrm{where}\quad a=\frac{\beta}{\alpha+\beta},\quad b=\frac{\alpha}{\alpha+\beta}
    \end{gathered}
\end{equation}

\noindent where $a$ and $b$ represent the model and data scaling exponents, respectively. In order to present $\alpha$ and $\beta$ with scaling exponents, we have

\begin{equation}
    \frac{\alpha}{\beta}=\frac{1}{a}-1
\end{equation}

Let $\eta \doteq \alpha + \beta$, the parametric loss $\hat{L}$ can be presented as:

\begin{equation}
    \begin{gathered}
        \hat{L} (N, D) = E + \frac{A}{N^{(1-a)\eta}} + \frac{B}{D^{a\eta}} \\
    \end{gathered}    
\end{equation}

Then, we can obtain the partial derivatives of $\hat{L}$ with respect to N expressed in terms of $a$ and $b$.

\begin{equation}
    \frac{\partial \hat{L}}{\partial N}=A\cdot (a-1)\eta \cdot N^{(a-1)\eta-1} 
\end{equation}

It's obvious that

\begin{equation}
    \frac{\partial \hat{L}}{\partial N} < 0
\end{equation}

\noindent which means that the expected loss decreases when model size increases under same training tokens.

We can further get 

\begin{equation}
    \begin{aligned}
        \frac{\partial^2 L}{\partial a \partial N} &= A\cdot \eta \cdot N^{(a-1)\eta-1}\\
        &\quad + A\cdot (a-1)\eta \cdot \eta\cdot \ln{N}\cdot N^{(a-1)\eta-1}\\
        &= A\cdot \eta \cdot N^{(a-1)\eta-1}\\
        &\quad \cdot \left[1 + (a-1)\eta\cdot \ln{N} \right]
    \end{aligned}
\end{equation}

Because $A, N, \alpha, \beta, \eta > 0$, we have

\begin{equation}
    A \cdot \eta \cdot N^{(a-1)\eta-1} > 0
\end{equation}

\noindent and since $1>a>0, \eta>0, N\gg1$, we have
 \begin{equation}
     1 + (a-1)\eta\cdot \ln{N} < 0
 \end{equation}

Thus, we have

\begin{equation}
    \frac{\partial^2 L}{\partial a \partial N} < 0
\end{equation}

That means that at a specific $N_0$, the slope of the tangent to the $\hat{L}-N$ curve ($\frac{\partial \hat{L}}{\partial N}$) decreases as $a$ increases (i.e., the larger the $a$, the steeper the tangent). In all, we've proven that

\begin{equation}
    \begin{gathered}
        \frac{\partial (-\hat{L})}{\partial N} > 0, \quad
        \frac{\partial^2 (-\hat{L})}{\partial a \partial N} > 0
    \end{gathered}
\end{equation}

Owing to this monotonic relationship, we can infer the value of $a$ from the slope of the tangent. For a given $N_0$, a steeper tangent (with a greater absolute value of the slope) indicates a larger $a$:

\begin{equation}
    a \propto -\frac{\partial L}{\partial N} \bigg|_{N = N_0}
\end{equation}

\begin{table*}[!htbp]
\centering
\caption{
Ablations on hyperparameters used in training 1.3B language models. Numbers in \gray{gray} represent the default values, as shown in Table~\ref{table:hyperpara-lm}. \\ \textbf{Abbreviations:} BS. = Batch Size, Hella. = HellaSwag, Winog. = Winogrande.
}
\label{tab:abla-hyper}
\vspace{1mm}
\begin{adjustbox}{max width=\textwidth}
    \begin{tabular}{lccccccccc}
    \hline
    Learning Rate & Global BS. & Hella. & LAMBADA & Winog. & PIQA & ARC & OpenbookQA & BoolQ & Avg\\ \hline
    \multicolumn{2}{l}{\textit{Binary Classification}} & & & & & & & & \\
    \gray{$2.5 \times 10^{-4}$} & \gray{256} & 48.13 & \textbf{48.96} & 54.30 & 69.75 & 41.66 & 30.40 & 61.38 & 50.65 \\
    \gray{$2.5 \times 10^{-4}$} & 512 & 46.63 & 47.72 & 53.51 & 68.34 & 40.81 & 31.20 & 59.85 & 49.72 \\
    $5.0 \times 10^{-4}$ & \gray{256} & 49.22 & 48.71 & 54.62 & 70.40 & 42.45 & \textbf{33.20} & 54.83 & 50.49 \\ \hline
    \textit{\OurMethod} & & & & & & & & & \\
    \gray{$2.5 \times 10^{-4}$} & \gray{256} & 49.07 & 48.42 & 55.09 & \textbf{70.57} & 42.67 & 31.40 & \textbf{61.68} & 51.27 \\
    \gray{$2.5 \times 10^{-4}$} & 512 & 46.51 & 46.24 & 52.33 & 69.37 & \textbf{44.84} & 30.00 & 61.56 & 50.12 \\
    $5.0 \times 10^{-4}$ & \gray{256} & \textbf{49.29} & 48.15 & \textbf{57.06} & 70.24 & 42.95 & 32.80 & 60.73 & \textbf{51.60} \\ \hline
    \end{tabular}
\end{adjustbox}
\end{table*}

\subsubsection{Generalizing from tangent slope to secant slope}
\label{app:dv2}
It's impossible to calculate the slope of the tangent $\frac{\partial L}{\partial N}$ in the real scenario, we can only acquire the slope of the secant line by assessing the cross-entropy loss on two models with different sizes (i.e. a pair of \textit{meta-models}). Next, we will prove that the slope of the tangent has a positive relationship with the slope of the secant line. Therefore, we can build direct relationship between the slope of the secant line and $a$.

Given a pair of meta-models with $N_p$ and $N_q$ parameters where $N_p < N_q$, we can denote the slope of the secant line as:

\begin{equation}
    \frac{\Delta \hat{L}}{\Delta N} = \frac{\hat{L}(N_q) - \hat{L}(N_p)}{N_q - N_p} = \frac{\int_{N_p}^{N_q} \frac{\partial L}{\partial N} dN}{N_q - N_p}
\end{equation}

For a larger $a$, $\frac{\partial L}{\partial N} \big|_{N = N_0} < 0$ is smaller for every $N_0$. This lead to a smaller $\frac{\Delta \hat{L}}{\Delta N}$, or a larger $-\frac{\Delta \hat{L}}{\Delta N}$, that is

\begin{equation}
    a \propto -\frac{\Delta \hat{L}}{\Delta N}
\end{equation}

\subsection{Sampling \textit{vs.} top-k selection} Previous works~\citep{gpt3, pile, dsir, qurating} typically use sampling without replacement, selecting data based on a rating score to balance quality and diversity. This approach often results in improved downstream performance. We conducted experiments to determine whether this sampling strategy could enhance the downstream performance of \OurMethod. Following~\citep{qurating}, we introduced a temperature term $\tau$ to adjust sample diversity. Here, $\tau \rightarrow 0$ means top-k selection, while $\tau \rightarrow \infty$ indicates uniform sampling. Results in Table~\ref{tab:sampling} indicate that top-k selection is the optimal data selection method for \OurMethod due to its reference-free nature and the unnecessary use of noisy sampling strategies to enhance data diversity.

\begin{table*}[!tbp]
\centering
\caption{
Ablations on sampling vs. top-k selection. Note that the top-k results are identical to the original \OurMethod results reported in Table~\ref{tab:vs-quality-filter}. We compare top-k data selection to sampling without replacement following~\citep{dsir, qurating}.
}
\label{tab:sampling}
\begin{adjustbox}{max width=\textwidth}
    \begin{tabular}{lcccccccc}
    \hline
    Sampling Method & Hellaswag & LAMBADA & Winogrande & PIQA & ARC & OpenbookQA & BoolQ & Avg \\ \hline
    $\tau = 0$ \ (Top-k) & \textbf{49.07} & \textbf{48.42} & \textbf{55.09} & \textbf{70.57} & \textbf{42.67} & 31.40 & \textbf{61.68} & \textbf{51.27} \\ 
    $\tau = 1$ & 47.99 & 45.93 & 53.91 & 68.50 & 41.46 & 31.80 & 61.13 & 50.10 \\ 
    $\tau = 2$ & 46.93 & 48.34 & 54.06 & 69.75 & 41.01 & \textbf{32.60} & 60.28 & 50.42 \\ 
    $\tau = 3$ & 47.14 & \textbf{48.42} & 54.46 & 70.02 & 42.28 & 32.00 & 59.82 & 50.59 \\  
    \hline
    \end{tabular}
\end{adjustbox}
\vspace{-3mm}
\end{table*}

\subsection{Ablation Study on Hyperparameters}

To further validate the robustness of \OurMethod, we conducted ablation experiments using various training hyperparameters on 1.3B models. Our focus was primarily on two hyperparameters: learning rate (default $2.5 \times 10^{-4}$) and global batch size (default 256). We doubled the default values for each in the ablation study. The results are presented in Table~\ref{tab:abla-hyper}. The results indicate that an increase in global batch size significantly reduces performance in both settings with different quality filters, as it halves the training steps. Conversely, increasing the learning rate slightly affects downstream accuracy. Overall, \OurMethod remains robust across a range of training hyperparameters, consistently surpassing binary classification, its top competitor, as shown in Table~\ref{tab:vs-quality-filter}.

\end{document}